 \documentclass[final,5p,times,twocolumn]{elsarticle}

\usepackage[english]{babel}
\usepackage{lineno}
\usepackage{amssymb, amsmath}
\usepackage{amsmath,amsfonts,amssymb,amsthm,epsfig,url,epstopdf,array}
\usepackage{graphicx}
\usepackage[justification=centering]{caption}
\usepackage{multirow}
\usepackage{hyperref}
\usepackage{enumerate}
\usepackage{color}
\usepackage[pdftex,dvipsnames]{xcolor}
\usepackage{float}
\usepackage{subcaption}
\usepackage[flushleft]{threeparttable}
\usepackage{booktabs,caption}
\usepackage{xargs}
\usepackage{pgf}
\usepackage{tikz}
\usepackage{relsize}
\usetikzlibrary{arrows,automata,positioning}
\usepackage[flushleft]{threeparttable} 
\usepackage{algorithm}
\usepackage{algpseudocode}

\hypersetup{pdfauthor={Name}}

\journal{Machine Learning with Applications}

\begin{document}
\begin{frontmatter}

\title{Long Short-term Cognitive Networks}


\author[TILBURG]{Gonzalo N\'apoles\corref{mycorrespondingauthor}}
\address[TILBURG]{Department of Cognitive Science \& Artificial Intelligence, Tilburg University, The Netherlands.}
\ead{g.r.napoles@uvt.nl}
\cortext[mycorrespondingauthor]{Corresponding author}

\author[TUE]{Isel Grau}
\address[TUE]{Information Systems Group, Eindhoven University of Technology, The Netherlands.}
\ead{i.d.c.grau.garcia@tue.nl}

\author[WARSAW]{Agnieszka Jastrzebska}
\address[WARSAW]{Faculty of Mathematics and Information Science, Warsaw University of Technology, Poland.}
\ead{A.Jastrzebska@mini.pw.edu.pl}

\author[TALCA]{Yamisleydi Salgueiro}
\address[TALCA]{Department of Computer Science, Faculty of Engineering, Universidad de Talca, Campus Curic\'o, Chile.}
\ead{yamisalgueiro@gmail.com}

\begin{abstract}
In this paper, we present a recurrent neural system named \textit{Long Short-term Cognitive Networks} (LSTCNs) as a generalization of the Short-term Cognitive Network (STCN) model. Such a generalization is motivated by the difficulty of forecasting very long time series efficiently. The LSTCN model can be defined as a collection of STCN blocks, each processing a specific time patch of the (multivariate) time series being modeled. In this neural ensemble, each block passes information to the subsequent one in the form of weight matrices representing the prior knowledge. As a second contribution, we propose a deterministic learning algorithm to compute the learnable weights while preserving the prior knowledge resulting from previous learning processes. As a third contribution, we introduce a feature influence score as a proxy to explain the forecasting process in multivariate time series. The simulations using three case studies show that our neural system reports small forecasting errors while being significantly faster than state-of-the-art recurrent models.
\end{abstract}


\end{frontmatter}


\section{Introduction}
\label{sec:introduction}

Time series analysis and forecasting techniques process data points that are ordered in a discrete-time sequence. While time series analysis focuses on extracting meaningful descriptive statistics of the data, time series forecasting uses a model for predicting the next value(s) of the series based on the previous ones. Traditionally, time series forecasting has been tackled with statistical techniques based on auto-regression or the moving average, such as Exponential Smoothing (ETS) \cite{hyndman2008forecasting} and the Auto-regressive Integrated Moving Average (ARIMA) \cite{box2015time}. These methods are relatively simple and perform well in univariate scenarios and with relatively small data.  However, they are more limited in predicting a long time horizon or dealing with multivariate scenarios.

The ubiquitousness of data generation in today's society brings the opportunity to exploit Recurrent Neural Network (RNN) architectures for time series forecasting. RNN-based models have reported promising results in multivariate forecasting of long series \cite{Hewamalage2021}. In contrast to feed-forward neural networks, RNN-based models capture long-term dependencies in the time sequence through their feedback loops. The majority of works published in this field are based on vanilla RNNs, Long-short Term Memory (LSTM) \cite{Hochreiter1997} or Gated Recurrent Unit (GRU) \cite{cho-etal-2014-learning} architectures. In the last M4 forecasting competition \cite{MAKRIDAKIS2018802}, the winners were models combining RNNs with traditional forecasting techniques, such as exponential smoothing \cite{smyl2020hybrid}. However, the use of RNN architectures is not entirely embraced by the forecasting community due to their lack of transparency, need for very specific configurations, and high computational cost \cite{Hewamalage2021,MAKRIDAKIS2018802,makridakis2018statistical}.

In this regard, the development of RNN architectures for time series forecasting can bring serious financial and environmental costs. As anecdotal evidence, one of the participants in the last M4 forecasting competition reported getting a huge electricity bill from 5 computers running for 4.5 months \cite{MAKRIDAKIS2018802}. More formally, the authors in \cite{Strubell_Ganesh_McCallum_2020}, presented an eye-opening study characterizing the energy required to train recent deep learning models, including their estimated carbon footprint. An example of a training-intensive task is the tuning of the BERT model \cite{devlin2018bert} for natural language processing tasks, which compares to the $CO_2$ emissions of a trans-American flight. One of the conclusions of the study in \cite{Strubell_Ganesh_McCallum_2020} is that researchers should focus on developing more efficient techniques and report measures (such as the training time) next to the model's accuracy. 

A second concern related to the use of deep machine learning models is their lack of interpretability. For most high-stakes decision problems having an accurate model is insufficient; some degree of interpretability is also needed. There exist several model-agnostic post-hoc methods for computing explanations based on the predictions of a black-box model. For example, feature attribution methods such as SHAP \cite{NIPS2017_7062} approximate the Shapley values that explain the role of the features in the prediction of a particular instance. Other techniques such as LIME \cite{ribeiro2016} leverage the intrinsic transparency of other machine learning models (e.g., linear regression) to approximate the decisions locally. In contrast, intrinsically interpretable methods provide explanations from their structure and can be mappable to the domain \cite{grau2020}. In \cite{Rudin2019}, the author argues that these explanations are more reliable and faithful to what the model computes. However, developing environmental-friendly RNN-based forecasting models able to provide a certain degree of transparency is a significant challenge. 

In this paper, we propose the Long Short-term Cognitive Networks (LSTCNs) to cope with the efficient and transparent forecasting of long univariate and multivariate time series. LSTCNs involve a sequential collection of Short-term Cognitive Network (STCN) blocks \cite{Napoles2019a}, each processing a specific time patch in the sequence. The STCN model allows for transparent reasoning since both weights and neurons map to specific features in the problem domain. Besides, STCNs allow for hybrid reasoning since the experts can inject knowledge into the network using prior knowledge matrices. As a second contribution, we propose a deterministic learning algorithm to compute the tunable parameters of each STCN block in a deterministic fashion. The highly efficient algorithm replaces the non-synaptic learning method presented in \cite{Napoles2019a}. As a final contribution, we present a feature influence score as a proxy to explain the reasoning process of our neural system. The numerical simulations using three case studies show that our model produces high-quality predictions with little computational effort. In short, we have found that our model can be remarkably faster than state-of-the-art recurrent neural networks.

The rest of this paper is organized as follows. Section~\ref{sec:literature} revises the literature on time series forecasting with recurrent neural networks, while Section~\ref{sec:stcn} presents the theory behind the STCN block. Section~\ref{sec:proposal} is devoted to LSTCN's architecture, learning, and interpretability. Section~\ref{sec:simulations} evaluates the performance of our model using three case studies involving long univariate and multivariate time series. Section~\ref{sec:remarks} concludes the paper and provides future research directions.

\section{Related work on time series forecasting}
\label{sec:literature}

In the last decade, we observed a constantly growing share of Artificial Neural Network-based approaches for time series forecasting. Prominent studies, including \cite{Bhaskar2012} and \cite{Ticknor2013}, use traditional feed-forward neural architectures trained with the backpropagation algorithm for time series prediction. However, in more recent papers, we see a shift towards other neural models. In particular, RNNs have gained momentum \cite{Kong2019}.

Feed-forward neural networks consist of layers of neurons that are one-way connected, from the input to the output layer, without cycles. In contrast, RNNs allow connections to previous layers and self-connections, resulting in cycles. In the special case of a fully connected Recurrent Neural Network \cite{Menguc2018}, the outputs of all neurons are also the inputs of all neurons. The literature is rich with various RNN architectures applied to time series forecasting. Yet, we can generalize the elaboration on various RNNs by stating that they allow having self-connected hidden layers \cite{Chen2019}.  Compared with Feedforward Neural Networks, RNNs utilize the action of hidden layer \textit{unfolding}, which makes them able to process sequential data. This explains their vast popularity in the analysis of temporal data, such as time series or natural language \cite{Cortez2018}.

A popular RNN architecture is called Long Short-Term Memory (LSTM). It was designed by \citet{Hochreiter1997} to overcome the problems arising when training vanilla RNN models. Traditional RNN training takes a very long time, mostly because of insufficient, decaying error when doing the error backpropagation \cite{Guo2017}. The LSTM architecture uses a special type of neurons called memory cells that mimic three kinds of gate operations \cite{Hewamalage2021}. These are referred to as the multiplicative input, output, and forget gates. These gates filter out unrelated and perturbed inputs \cite{Gao2019}. Standard LSTM models are constructed in a way that past observations influence only the future ones, but there exists a variant called bidirectional LSTM that lifts this restriction \cite{Cui2020}. Numerous studies show that both unidirectional and bidirectional LSTM networks outperform traditional RNNs due to their ability to capture long-term dependencies more precisely \cite{Tang2019}.  

The Gated Recurrent Unit (GRU) is another RNN model \cite{cho-etal-2014-learning}. In comparison to LSTM, GRU executes simplified gate operations by using only two types of memory cells: input merged with output and forget cell \cite{Wang2018}, called here update and reset gate, respectively \cite{Becerra2020}. As in the case of LSTM, GRU training is less sensitive to the vanishing/exploding gradient problem that is encountered in traditional RNNs \cite{Ding2019}. 

The inclusion of recurrent/delayed connections boosted the capability of neural models to predict time series accurately, while further improvements of their architecture (like LSTM or GRU model) made training dependable. However, it shall be mentioned that the use of the traditional error backpropagation is not the only option to learn network weights from historical data. Alternatively, we can use meta-heuristic approaches to train a model. There exists a range of interesting studies, where the authors used Genetic Algorithm \cite{Sadeghi2020} or Ant Colony Optimization \cite{ElSaid2018}. The study in \cite{Abdulkarim2019} concluded that, for the tested time series, dynamic Particle Swarm Optimization obtained a similar forecasting error compared with a feed-forward neural architecture and a recurrent one. 

It shall be noted that the application of a modern neural architecture does not relieve a model designer from introducing required data staging techniques. This is why we find a~range of domain-dependent studies that link various RNN architectures with supplemental processing options. For example, \citet{Liu2020} used a wavelet transform alongside a~GRU architecture, while \citet{Nikolaev2019} included a~regime-switching step, and \citet{Cheng2018} employed wavelet-based de-noising and adaptive neuro-fuzzy inference system.

We should mention recent studies on fusing RNN architectures with Convolutional Neural Networks (CNNs). The latter model has attracted much attention due to its superior efficiency in pattern classification. We find a range of studies \cite{Tanwi2020,Xue2019}, where a CNN is merged with an RNN in a deep neural model that aims at time series forecasting. The role of a CNN is to extract features that are used to train an RNN forecasting model \cite{Li2020}. Attention mechanisms have also been successfully merged with RNNs, as presented by \citet{Zhang2020}.

From a high-level perspective on time series forecasting with RNNs, we can also distinguish architectures that read in an entire time series and produce an internal representation of the series, i.e., a network plays the role of an encoder \cite{Laubscher2019}. A decoder network then needs to be used to employ this internal representation to produce forecasts \cite{Bappy2019}. The described scheme is called an encoder-decoder network and was applied, for example, by \citet{Habler2018} together with LSTM, by \citet{Chen2020}  with convolutional LSTM, and by \citet{Yang2021} with GRU.

We shall also mention the forecasting models based on Fuzzy Cognitive Maps (FCMs) \cite{Kosko1986}. Such networks are knowledge-oriented architectures with processing capabilities that mimic the ones of RNNs. The most attractive feature of these models is network interpretability. There are numerous papers, including the works of \citet{Papageorgiou2017}, \citet{Vanhoenshoven2020}, \citet{Pedrycz2016} or \citet{Wu2017}, where FCMs are applied to process temporal data. However, recent studies show that even better forecasting capabilities can be achieved with STCNs \cite{Napoles2019a} or Long-term Cognitive Networks \cite{Napoles2020a}. As far as we know, these FCM generalizations have not yet been used for time series forecasting. This paper extends the research on the STCN model, which will be used as the main building block of our proposal.

\section{Short-term Cognitive Networks}
\label{sec:stcn}

The STCN model was introduced in \cite{Napoles2019a} to cope with short-term WHAT-IF simulation problems where problem variables are mapped to neural concepts. In these problems, the goal is to compute the immediate effect of input variables on output ones given long-term prior knowledge. Remark that the model in \cite{Napoles2019a} was trained using a gradient-based non-synaptic learning approach devoted to adjusting a set of parametric transfer functions. In this section, we redefine the STCN model such that it can be trained in a synaptic fashion.

The STCN block involves four matrices to perform reasoning: $W_1^{(t)}$, $B_1^{(t)}$, $W_2^{(t)}$, and $B_2^{(t)}$. The first two matrices denote the prior knowledge coming from a previous learning process and can be modified by the experts to include new pieces of knowledge that have not yet been recorded in the historical data (e.g., an expected increase in the Bitcoin value as Tesla decides to accept such cryptocurrency as a valid payment method). These prior knowledge matrices allow for hybrid reasoning, which is an appealing feature of the STCN model. The third and fourth matrices contain learnable weights that adapt the input $X^{(t)}$ and the prior knowledge to the expected output $Y^{(t)}$ in the current step. The matrices $B_1^{(t)}$ and $B_2^{(t)}$ represent the bias weights. 

\begin{figure}[!ht]
\centering
	\includegraphics[width=0.5\textwidth]{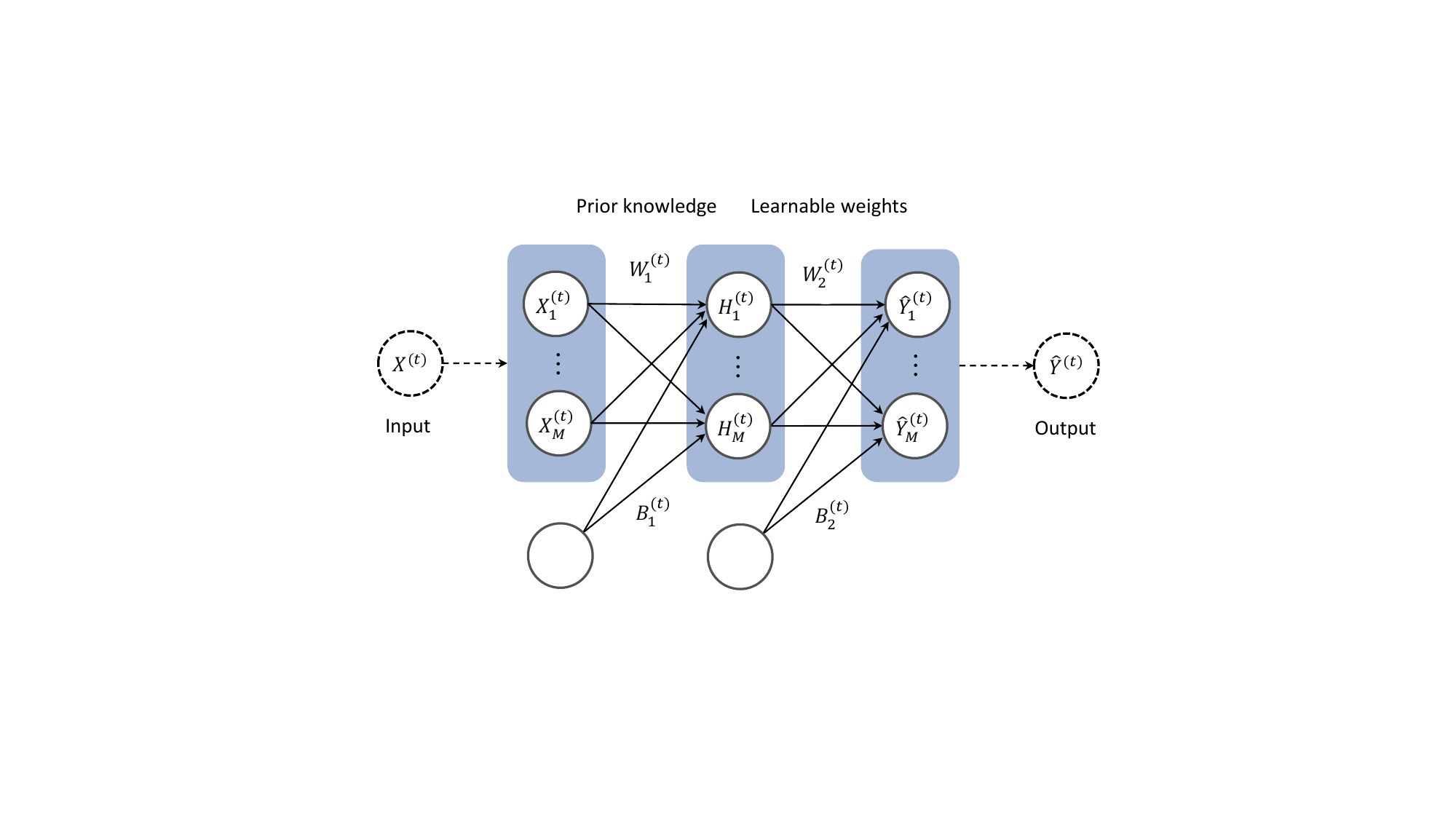}
\caption{The STCN block involves two components: the prior knowledge matrices $W^{(t)}_1$ and $B^{(t)}_1$, and the learnable matrices $W^{(t)}_2$ and $B^{(t)}_2$. The prior knowledge matrices are a result of a previous learning process and can be modified by domain experts if deemed opportune.}
\label{fig:network}
\end{figure}

Figure \ref{fig:network} shows how the different components interact with each other in an STCN block. It is important to highlight that this model lacks hidden neurons, so each inner block (abstract layer) has exactly $M$ neurons, with $M$ being the number of neural concepts in the model. This means that we have a neural system in which each component has a well-defined meaning. For example, the intermediate state $H^{(t)}$ represents the outcome that the network would have produced given $X^{(t)}$ if the network would not have been adjusted to the expected output $Y^{(t)}$. Similarly, the bias weights denote the external information that cannot be inferred from the given inputs.

Equations \eqref{eq:stcn1} and \eqref{eq:stcn2} formalize the short-term reasoning process of this model in the $t$-th iteration,

\begin{equation}
\label{eq:stcn1}
\hat{Y}^{(t)}=f\left(H^{(t)} W_2^{(t)} \oplus B_2^{(t)}\right)
\end{equation}
\noindent and
\begin{equation}
\label{eq:stcn2}
H^{(t)}=f\left(X^{(t)} W_1^{(t)} \oplus B_1^{(t)} \right)
\end{equation}

\noindent where $X^{(t)}$ and $\hat{Y}^{(t)}$ are $K \times M$ matrices encoding the input and the forecasting in the current iteration, respectively, with $K$ being the number of observations and $M$ the number of neurons. $B_{1}$ and $B_{2}$ are ${1\times M}$ matrices representing the bias weights. In these equations, the $\oplus$ operator performs a matrix-vector addition by operating each row of a given matrix with a vector, provided that both the matrix and the vector have the same number of columns. Finally, $f(\cdot)$ stands for the non-linear transfer function, typically the sigmoid function:

\begin{equation}
\label{eq:sigmoid}
f(x) = \frac{1}{1+e^{-x}}.
\end{equation}

The inner working of an STCN block can be summarized as follows. The block receives a weight matrix $W_1^{(t)}$, the bias weight matrix $B_1^{(t)}$ and a chunk of data $X^{(t)}$ as the input data. Firstly, we compute an intermediate state $H^{(t)}$ that mixes $X^{(t)}$ with the prior knowledge (e.g., knowledge resulting from the previous iteration). Secondly, we operate $H^{(t)}$ with $W_2^{(t)}$ and $B_2^{(t)}$ to approximate the expected output $Y^{(t)}$.

This short-term reasoning of this model makes it less sensitive to the convergence issues of long-term cognitive networks such as the unique-fixed point attractors \cite{Napoles2019a}. Furthermore, the short-term reasoning allows extracting more clear patterns to be used to generate explanations.

\section{Long Short-term Cognitive Network}
\label{sec:proposal}

In this section, we introduce the \textit{Long Short-term Cognitive Networks} for time series forecasting, which can be defined as a collection of chained STCN blocks.

\subsection{Architecture}
\label{sec:proposal:model}

As mentioned, the model presented in this section is devoted to the multiple-ahead forecast of very long (multivariate) time series. Therefore, the first step is splitting the time series into $T$ time patches, each comprising a collection of tuples with the form $(X^{(t)},X^{(t+1)})$. In these tuples, the first matrix denotes the input used to feed the network in the current iteration, while the second one is the expected output $Y^{(t)}=X^{(t+1)}$. Notice that each time patch often contains several time steps (e.g., all tuples produced within a 24-hour time frame).

\begin{figure}[!ht]
\centering
	\includegraphics[width=0.48\textwidth]{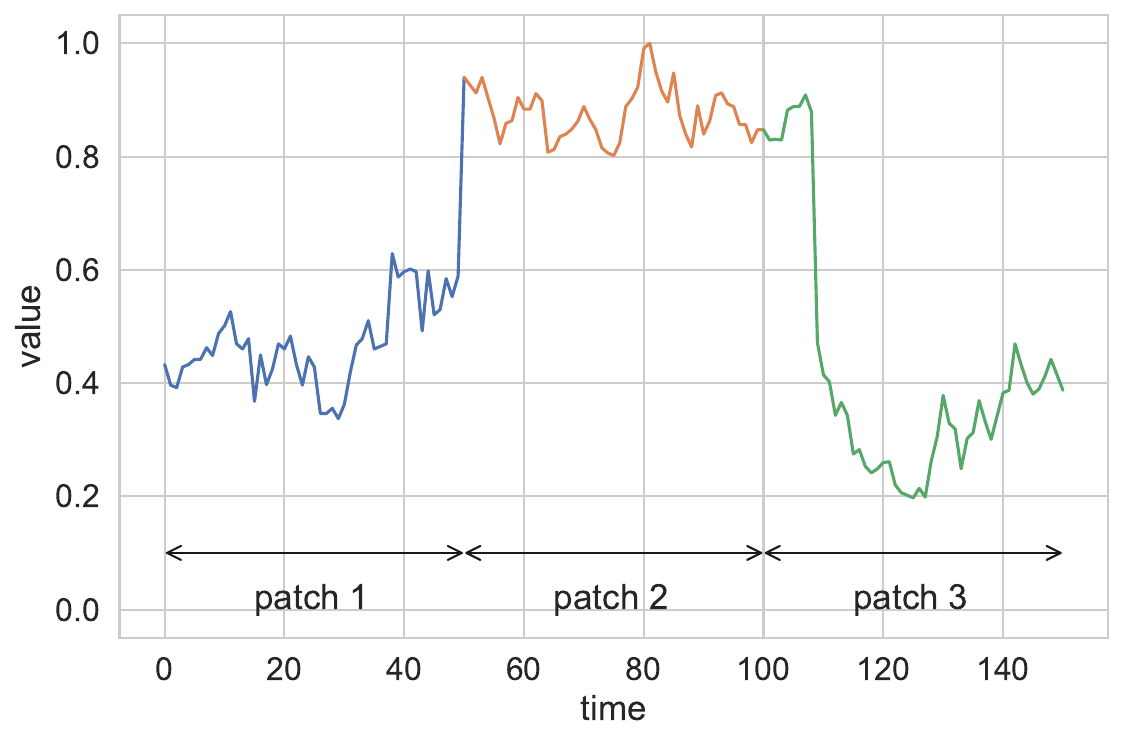}
\caption{Recurrent approach to process a (multivariate) time series with an LSTCN model. The sequence is split into $T$ time patches with even length. Each time patch is used to train an STCN block that employs information from the previous block as prior knowledge.}
\label{fig:timepatches}
\end{figure}

Figure \ref{fig:timepatches} shows, as an example, how to decompose a given time series into $T$ time patches of equal length where each time patch will be processed by an STCN block. This procedure holds for multivariate time series such that both $X^{(t)}$ and $Y^{(t)}$ have a dimension of $K \times M$. In this case, $K$ denotes the number of time steps allocated to the time patch, whereas $M$ defines the width of each STCN block. Therefore, if we have a multivariate time series described by $N$ continuous variables and want to forecast $L$ steps, then $M=N \times L$.

In short, the LSTCN model can be defined as a collection of STCN blocks, each processing a specific time patch and passing knowledge to the next block. In each time patch, the matrices of the previous model are aggregated and used as prior knowledge for the current STCN block, that is to say:

\begin{equation}
\label{eq:aggregation1}
W_1^{(t)}=\Psi(W_1^{(t-1)},W_2^{(t-1)})
\end{equation}
\noindent and
\begin{equation}
\label{eq:aggregation2}
B_1^{(t)}=\Psi(B_1^{(t-1)},B_2^{(t-1)})
\end{equation}

\noindent such that $\Psi(x,y)=tanh(max\{x,y\})$. The aggregation procedure creates a chained neural structure that allows for long-term predictions since the learned knowledge is used when performing reasoning in the current iteration.

\begin{figure}[!ht]
\centering
	\includegraphics[width=0.48\textwidth]{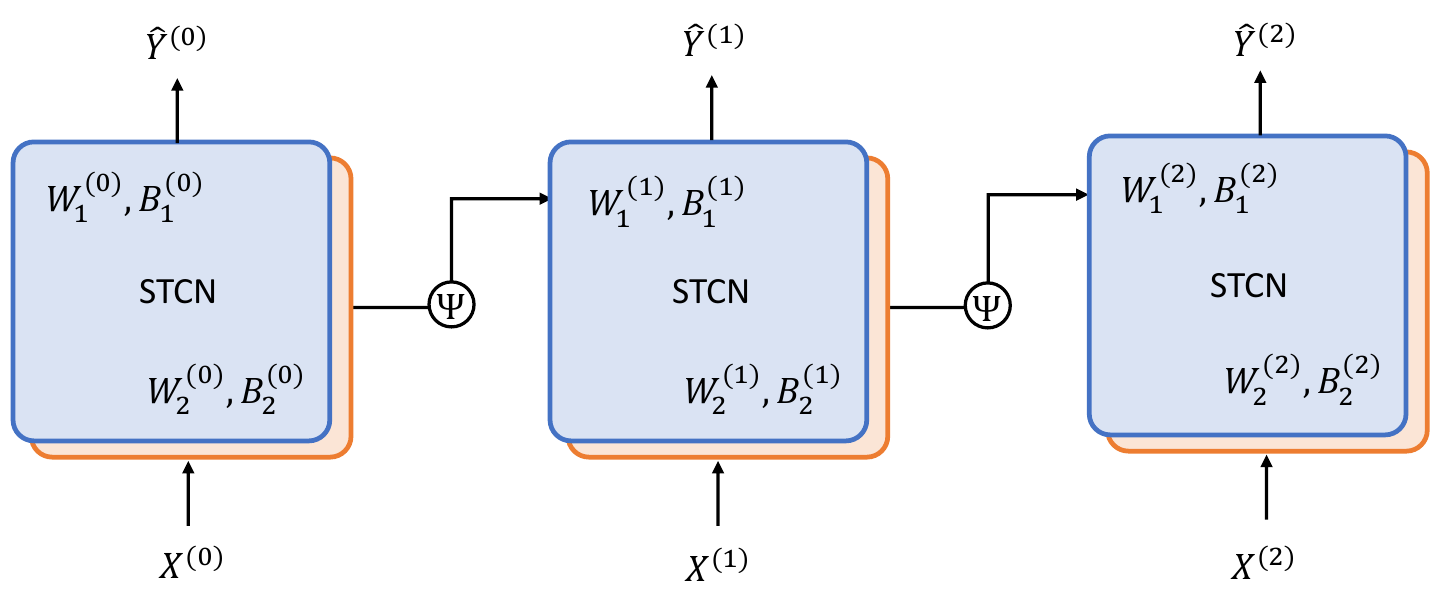}
\caption{Example of an LSTCN composed of three STCN blocks. In each iteration, the model receives a time patch $X^{(t)}$ to be processed and produces an approximation of the expected output $Y^{(X)}$. The weights learned in the current block are aggregated (using Equations \eqref{eq:aggregation1} and \eqref{eq:aggregation2} and transferred to the following STCN block as prior knowledge matrices.}
\label{fig:lstcn}
\end{figure}

Figure \ref{fig:lstcn} shows the LSTCN architecture to process the time series in Figure \ref{fig:timepatches}, which was split into three time patches of equal length. In the figure, blue boxes represent STCN blocks, while orange boxes denote learning processes.

It should be highlighted that, although the LSTCN model works in a sequential fashion, each STCN block performs an independent learning process (to be explained in the next subsection) before moving to the next block. Therefore, the long-term component refers to how we process the whole sequence, which is done by transferring the knowledge (in the form of weights) from one STCN block to another. Notice that we do not pass the neurons' activation values to the subsequent blocks. Once we have processed the whole sequence, the model narrows down to the last STCN in the pipeline.

\subsection{Learning}
\label{sec:proposal:learning}

Training an LSTCN model means training each STCN block with its corresponding time patch. In this neural system, the learned knowledge up to the current iteration is stored in $B_1$ and $W_1$, while $B_2$ and $W_2$ contain the knowledge needed to make the prediction in the current iteration. Therefore, the learning problem consist of computing $W_2^{(t)}$ and $B_2^{(t)}$ given the tuple $(X^{(t)}, Y^{(t)})$ corresponding to the current time patch. The underlying optimization problem is given below:

\begin{equation}
\label{eq:learning1}
min \rightarrow \left\Vert f\left(H^{(t)} W_2^{(t)} \oplus B_2^{(t)}\right)- Y^{(t)}\right\Vert_{\ell_2} + \lambda \left\Vert \Gamma_2^{(t)} \right\Vert_{\ell_2}
\end{equation}
\noindent such that 
\begin{equation}
\label{eq:learning2}
\Gamma_2^{(t)} = \begin{bmatrix} W_2^{(t)} \\ B_2^{(t)} \end{bmatrix}
\end{equation}

\noindent represents the matrix that results after performing a row-wise concatenation of the bias weight matrix $B_2^{(t)}$ to $W_2^{(t)}$, while $\lambda \geq 0$ is the ridge regularization penalty. The added value of using a ridge regression approach is regularizing the model and preventing overfitting. In our network, overfitting is likely to happen when splitting the original time series into too many time patches covering few observations.

Equation \eqref{eq:ridge} displays the deterministic learning rule solving this ridge regression problem,

\begin{equation}
\label{eq:ridge}
\Gamma_2^{(t)} = \left( \left( \Phi^{(t)} \right)^{\top} \Phi^{(t)} + \lambda \Omega^{(t)} \right)^{-1} \left( \Phi^{(t)} \right)^{\top} f^{-} \left(Y^{(t)}\right)
\end{equation}

\noindent where $\Phi^{(t)}=(H^{(t)}|A)$ such that $A_{K \times 1}$ is a column vector filled with ones, while $\Omega^{(t)}$ denotes the diagonal matrix of $(\Phi^{(t)})^{\top} \Phi^{(t)}$. Remark that this learning rule assumes that the activation values in the inner layer are standardized. As far as standardization is concerned, these calculations are based on standardized activation values. When the final weights are returned, they are adjusted back into their original scale.

It should be mentioned that an STCN block trained using the learning rule in Equation \eqref{eq:ridge} is similar to an Extreme Learning Machine (ELM) \cite{Huang2006}. There are two main differences, though. Firstly, the $W_{1}^{(t)}$ and $B_{1}^{(t)}$ matrices are not random but initialized with the prior knowledge arriving at the STCN block from previous learning processes. Secondly, while the hidden layer of ELMs is of arbitrary width, the number of neurons in an STCN is given by the number of steps ahead to be predicted and the number of features in the multivariate time series. Hence, each neuron (also referred to as neural concept) represents the state of a problem feature in a given time step. While this constraint equips our model with interpretability features, it might also limit its approximation capabilities. 

Another issue that deserves attention is how to estimate the first weight matrix $W_1^{(0)}$ to be used as prior knowledge in the first iteration. This matrix is expected to be (partially) provided by domain experts or computed from a previous learning process (e.g., using a transfer learning approach). In this paper, we simulate such knowledge by fitting a stateless STCN (that is to say, $H^{(t)} = X^{(t)} $) on a smoothed representation of the whole time series we are processing. The smoothed time series is obtained using the moving average method for a given window size. Finally, we generate some white noise over the computed weights to compensate for the moving average operation. Equation \eqref{eq:init} shows how to compute this matrix,

\begin{equation}
\label{eq:init}
W_1^{(0)} \sim\mathcal{N}\left(\left( \bar{X}^{\top} \bar{X} + \lambda \Omega \right)^{-1} \bar{X}^{\top} f^{-} \left( \bar{Y} \right), \sigma \right)
\end{equation}

\noindent where $\bar{X}$ and $\bar{Y}$ are the smoothed inputs and outputs obtained for the whole time series, respectively, while $\sigma$ is the standard deviation. In this case, we will use $\Omega$ again to denote the diagonal matrix of $\bar{X}^{\top} \bar{X}$ if no confusion arises. 

The prior bias matrix $B_1^{(0)}$ is assumed to be zero since we use that component to model the external stimulus of neurons after performing an STCN's learning process.

The intuition dictates that the training error will go down as more time patches are processed. Of course, such time patches should not be too small to avoid overfitting. In some cases, we might obtain an optimal performance using a single time patch containing the whole sequence such that we will have a single STCN block. In other cases, it might occur that we do not have access to the whole sequence (e.g., as happens when solving online learning problems), such that using a single STCN block would not be an option.

\subsection{Interpretability}
\label{sec:proposal:measure}

As mentioned, the architecture of our neural system allows explaining the forecasting since both neurons and weights have a precise meaning for the problem domain being modeled. However, the interpretability cannot be confined to the absence of hidden components in the network since the graph might involve hundreds or thousands of edges. 

In this subsection, we introduce a measure to quantify the influence of each feature in the forecasting of \textit{multivariate} time series. The proposed measure can be computed from $W_1^{(t)}$, $W_2^{(t)}$ or their combination. The scores obtained from $W_1^{(t)}$ can be understood as the feature influence up to the $t$-th time patch, while scores obtained from $W_2^{(t)}$ can be understood as the feature influence to the current time patch.

Let us recall that $W_1^{(t)}$ and $W_2^{(t)}$ are $M \times M$ matrices such that $M=N \times L$, assuming that we have a multivariate time series with $N$ features and that we want to forecast $L$ steps ahead. Moreover, the neurons are organized temporally, which means that we have $L$ blocks of neurons, each containing $N$ units. Equations \eqref{eq:score1} and \eqref{eq:score2} show how to quantify the effect of feature $f_i$ on feature $f_j$ given a matrix $W^{(t)}$ that characterizes the interaction among the problem features,

\begin{equation}
\label{eq:score1}
\gamma^{(t)}(f_i,f_j) = \sum_{p_i \in P(i)} \sum_{p_j \in P(j)}  \left\vert w^{(t)}_{p_i p_j} \right\vert, w^{(t)}_{p_i p_j} \in W^{(t)}
\end{equation}

\noindent such that

\begin{equation}
\label{eq:score2}
P(i)= \{ p \in \mathbb{N}, p \leq M~|~(p~mod~i)=0\}.
\end{equation}

The \textit{feature influence score} in Equation \eqref{eq:score1} can be normalized such that the sum of all scores related to the $j$-th feature is one. This can be done as follows:

\begin{equation}
\label{eq:score3}
\hat{\gamma}^{(t)}(f_i,f_j) = \frac{\gamma^{(t)}(f_i,f_j)}{\sum_{k=1}^{N} \gamma^{(t)}(f_k,f_j)}.
\end{equation}

The rationale behind the proposed feature influence score is that the most important problem features will have attached weights with large absolute values. Moreover, it is expected for the learning algorithm to produce sparse weights with a zero-mean normal distribution, which is an appreciated characteristic when it comes to interpretability.

The idea of computing the relevance of features from the weights in neural systems has been explored in the literature. For example, the Layer-Wise Relevance Propagation (LRP) algorithm \cite{Bach2015} explains the predictions made by a neural classifier for a given instance by assigning relevance scores to features, which are computed using the learned weights and neurons' activation values. It should be stated that we do not use neurons' activation values in our feature influence score as we intend to produce global explanations based on the learned weights only. Similar approaches have been proposed in \cite{Napoles2021b} and \cite{Napoles2021c} but applied to LTCN-based classifiers.

\section{Numerical simulations}
\label{sec:simulations}

In this section, we will explore the performance (forecasting error and training time) of our neural system on three case studies involving univariate and multivariate time series. In the case of multivariate time series, we will also depict the feature contribution score to explain the predictions.

When it comes to the pre-processing steps, we interpolate the missing values (whenever applicable) and normalize the series using the \textit{min-max} method. In addition, we split the series into 80\% for training and validation and 20\% for testing purposes. As for the performance metric, we use the \textit{mean absolute error} in all simulations reported in the section. For the sake of convenience, we trimmed the training sequence (by deleting the first observations) such that the number of times is a multiple of $L$ (the number of steps ahead we want to forecast).

The models used for comparison are a fully connected Recurrent Neural Network (RNN) where the output is to be fed back to the input, GRU and LSTM. In these models, the number of epochs was set to 20, while the batch size was obtained through hyperparameter tuning (using grid search). The candidate batch sizes were the powers of two, starting from 32 until 4,096. The values for the remaining parameters were retained as provided in the Keras library. In the case of the LSTCN model, we fine-tuned the number of time patches $T \in \{1,2\ldots,10\}$ and the regularization penalty $\lambda \in \{\text{1.0E-3}, $ $ \text{1.0E-2}, $ $\text{1.0E-1}, \text{1.0E+1}, \text{1.0E+2}, \text{1.0E+3}\}$. In Equation \eqref{eq:init}, we arbitrarily set the standard deviation $\sigma$ to 0.05 and the moving window size $w$ to 100. These two parameters were not optimized during the hyperparameter tuning step as they were used to simulate the prior knowledge component.

Finally, all experiments presented in this section were performed on a high-performance computing environment that uses two Intel Xeon Gold 6152 CPUs at 2.10 GHz, each with 22 cores and 768 GB of memory.

\subsection{Apple Health's step counter}
\label{sec:simulations:steps}

The first case study concerns physical activity prediction based on daily step counts. In this case study, the health data of one individual was extracted from the Apple Health application in the period from 2015 to 2021. In total, the time series dataset is composed of 79,142 instances or time steps. The Apple Health application records the step counts in small sessions during which the walking occurs. The dataset\footnote{\url{https://bit.ly/2S9vzMD}} contains two timestamps (start date and end date), the number of recorded steps, and separate columns for year, month, date, day, and hour. Besides, the day of the week that each value was recorded is known. Table~\ref{tab:stats_steps} presents descriptive statistics attached to this univariate time series before normalization. 

\begin{table}[!htp]
\centering
\caption{Descriptive statistics for the Steps case study. }
\label{tab:stats_steps}
\begin{tabular}{|c|c|c|c|c|}
\hline
  variable     &mean&std &min&max	\\\hline
  value&191.63&	235.73&1.00&7,205.00\\\hline
\end{tabular}
\end{table}

The target variable (number of steps) follows an exponential distribution with very infrequent, extremely high step counts and very common low step counts. Overall, the data neither follows seasonal patterns nor a trend.




Table \ref{tab:steps} shows the normalized errors attached to the recurrent neural networks when forecasting 50 steps ahead in the Steps dataset. In addition, we portray the training and test times (in seconds) for the optimized models. The hyperparameter tuning reported that our neural system needed two iterations to produce the lowest forecasting errors, while the optimal batch size for RNN, GRU and LSTM was found to be 32. Although LSTCN outperforms the remaining methods in terms of forecasting error, what is truly remarkable is its efficiency. The results show that LSTCN is 2.7E+2 times faster than RNN, 2.3E+3 times faster than GRU, and 2.2E+3 times faster than LSTM. In this experiment, we ran all models five times with optimized hyperparameters and selected the shortest training time in each case. Hence, the time measures reported in Table \ref{tab:steps} concern the fastest executions observed in our simulations.

\begin{table}[!htp]
\centering
\caption{Simulation results for the Steps case study.}
\label{tab:steps}
\begin{tabular}{|c|c|c|c|c|}
\hline
       & \multicolumn{2}{c|}{error} & \multicolumn{2}{c|}{time} \\ \hline
method & training      & test       & training     & test       \\ \hline
RNN    & 0.0241        & 0.0249     & 5.513        & 0.444     \\ \hline
GRU    & 0.0236        & 0.0251     & 46.002       & 0.703      \\ \hline
LSTM   & 0.0237        & 0.0248     & 44.689       & 0.933      \\ \hline
LSTCN  & 0.0221        & 0.0212     & 0.0202       & 0.001       \\ \hline
\end{tabular}
\end{table}

\begin{figure}[!ht]
\centering
	\includegraphics[width=0.48\textwidth]{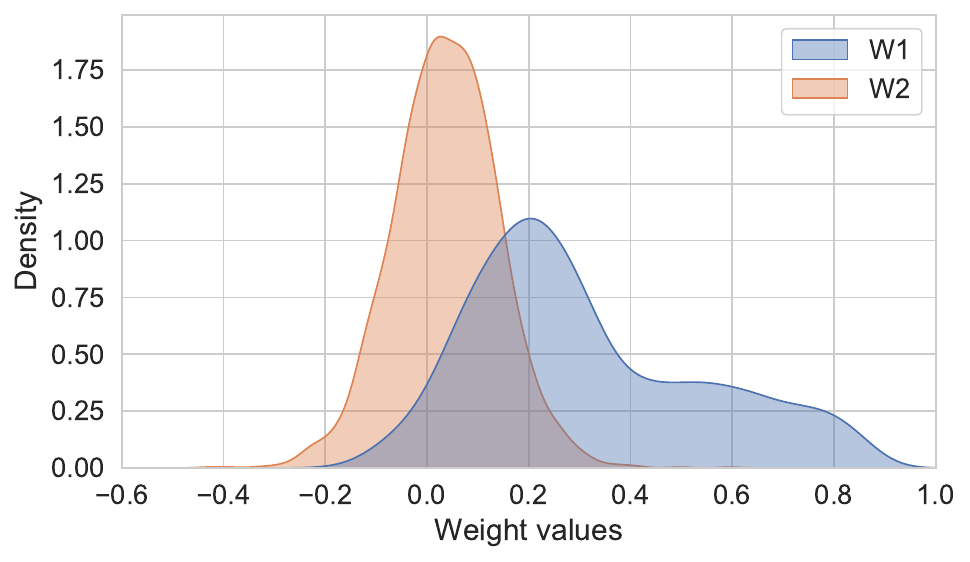}
\caption{Distribution of weights for the Steps case study.}
\label{fig:histogram_steps}
\end{figure}

Figure \ref{fig:histogram_steps} displays the distributions of weights in the $W_1$ and $W_2$ matrices attached to the last STCN block (the one to be used to perform the forecasting). In this case study, most prior knowledge weights are distributed in the $[-0.2,1.0]$ interval, while weights in $W_2$ follows a zero-mean Gaussian distribution. This figure illustrates that the network adapts the prior knowledge to the last piece of data available.  

\begin{figure}[!ht]
\centering
	\includegraphics[width=0.48\textwidth]{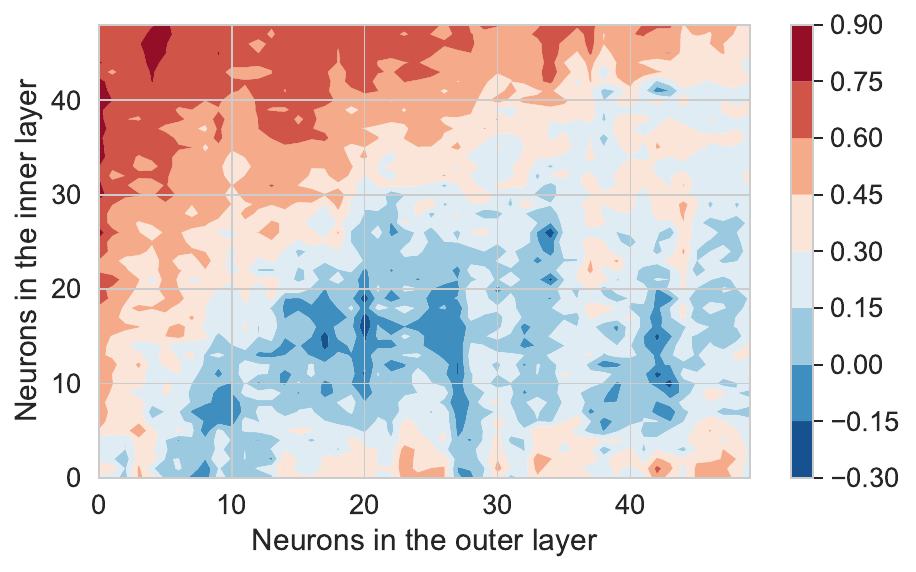}
\caption{Behavior of weights connecting the inner neurons with the outer ones in the last STCN block.}
\label{fig:contour_steps}
\end{figure}

Figure \ref{fig:contour_steps} depicts the overall behavior of weights connecting the inner neurons with the outer ones in the last STCN block. In this simulation, we averaged the $W_1$ and $W_2$ matrices for the sake of simplicity. Observe that the learning algorithm assigns larger weights to connections between neurons processing the last steps in the input sequence and neurons processing the first steps in the output sequence. This is an expected behavior in time series forecasting that supports the rationale of the proposed feature relevance measure.

The fact that each neuron has a well-defined meaning for the problem domain makes it possible to elucidate how the network uses the current $L$ time steps to predict the following ones. Using that knowledge, experts could estimate how many previous time steps would be needed to predict a sequence of length $L$ without performing further simulations. 

\subsection{Household electric power consumption}
\label{sec:simulations:power}

The second case study concerns the energy consumption in one house in France measured each minute from December 2006 to November 2010 (47 months). This dataset\footnote{\url{https://archive.ics.uci.edu/ml/datasets/individual+household+electric+power+consumption}} involves nine features and 2,075,259 observations from which 1.25\% are missing. Hence, records with missing values were interpolated using the nearest neighbor method. In our experiments, we retained the following variables: global minute-averaged active power (in kilowatt), global minute-averaged reactive power (in kilowatt), minute-averaged voltage (in volt), and global minute-averaged current intensity (in ampere).

\begin{table}[!htp]
\centering
\caption{Descriptive statistics concerning four variables in the Power case study: mean value, standard deviation, minimal, and maximal values (``pwr'' stands for power).}
\label{tab:stats_power}
\begin{tabular}{|c|c|c|c|c|}
\hline
  variable     &mean&std &min&max	\\\hline
  active pwr&	1.09&	1.06&		0.08&11.12\\\hline
  reactive pwr & 0.12&0.11&0.00&	1.39\\\hline
  voltage &240.84	&	3.24&	223.20&		254.15\\\hline
  intensity&4.63&	4.44&	0.20&	48.40\\\hline
\end{tabular}
\end{table}

The series exhibits cyclic patterns. On the most fine-grained scale, we observe a repeating low nighttime power consumption. We also noted a less distinct but still present pattern related to the day of the week: higher power consumption during weekend days. Finally, we observed high power consumption during the winter months (peaks in January) each year and low in summer (lowest values recorded for July).  



Table \ref{tab:power} portrays the normalized errors obtained by each optimized model when forecasting 200 steps ahead, and the training and test times (in seconds). The hyperparameter tuning reported that our network produced the optimal forecasts with two iterations, while the optimal batch size for RNN, GRU and LSTM was 4,096, 64 and 256, respectively. According to these simulations, LSTCN obtains the best results followed by GRU with the latter being notably slower than the former. Overall, LSTCN proved to be 2.3E+1 times faster than RNN, 2.2E+3 times faster than GRU, and 1.6E+3 times faster than LSTM. RNN was the second-fastest algorithm, although it showed the worst forecasting errors in our simulations.

\begin{table}[!htp]
\centering
\caption{Simulation results for the Power case study.}
\label{tab:power}
\begin{tabular}{|c|c|c|c|c|}
\hline
       & \multicolumn{2}{c|}{error} & \multicolumn{2}{c|}{time} \\ \hline
method & training   & test      & training  & test      \\ \hline
RNN    & 0.3326    & 0.3359    & 14.56     & 0.76      \\ \hline
GRU    & 0.0581    & 0.0547    & 1373.58   & 2.21      \\ \hline
LSTM   & 0.0637    & 0.0581    & 991.51    & 2.81     \\ \hline
LSTCN  & 0.0559    & 0.0531    & 0.63      & 0.04      \\ \hline
\end{tabular}
\end{table}

\begin{figure}[!ht]
\centering
	\includegraphics[width=0.48\textwidth]{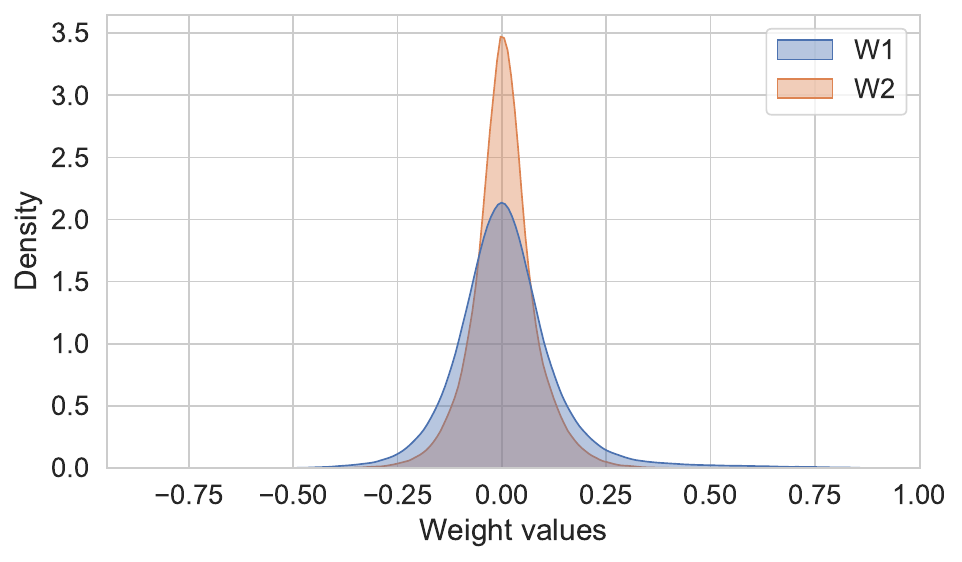}
\caption{Distribution of weights for the Power case study.}
\label{fig:histogram_power}
\end{figure}

Figure \ref{fig:histogram_power} displays the distributions of weights in the $W_1$ and $W_2$ matrices for the last STCN block. These histograms reveal that weights follow a zero-mean Gaussian distribution and that the second matrix has more weights near zero (the shape of the second curve contracts toward zero). In this case study, the network does not shift the distribution of weights as happened in the first case study. Actually, the accumulated prior knowledge does not seem to suffer much distortion (distribution-wise) when adapted to the last time patch.

\begin{figure}[!ht]
\centering
	\includegraphics[width=0.48\textwidth]{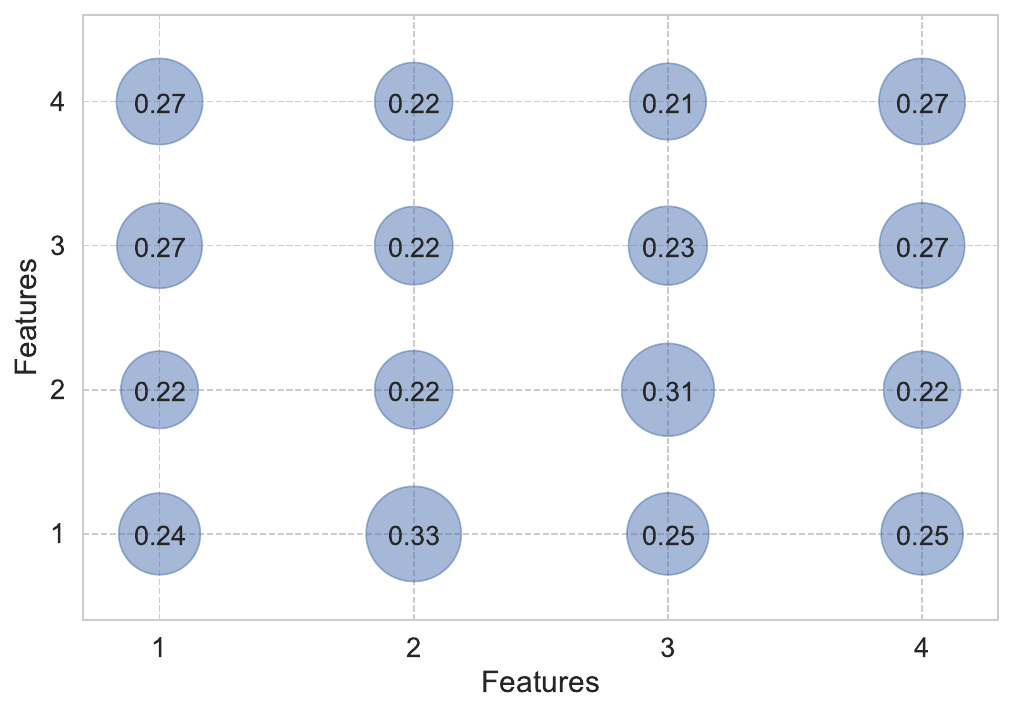}
\caption{Feature influence in the Power case study.}
\label{fig:relevance_power}
\end{figure}

Figure \ref{fig:relevance_power} displays the feature influence scores obtained with Equation \eqref{eq:score3}. These scores were computed after averaging the $W_1$ and $W_2$ matrices that result from adjusting the network to the last time patch. In this figure, the bubble size denotes the extent to which one feature in the y-axis is deemed relevant to forecast the value of another feature in the x-axis. For example, it was observed that the first feature (global active power) is the most important one to forecast the second feature (global reactive power). Observe that the sum of all scores by column is one due to the normalization step.

Overall, the results indicate that the LSTCNs obtain small forecasting errors while being markedly faster than the state-of-the-art recurrent neural networks. Moreover, the LSTCNs' knowledge structures facilitate explaining how the forecasting was made, using the model's feature relevance explanations.

\subsection{Bitcoin transaction graph change analysis}
\label{sec:simulations:c2}

In this section, we inspect a case study concerning changes in the Bitcoin transaction graph observed with a daily frequency from January 2009 to December 2018. The data set is publicly available in the UCI Repository\footnote{\url{https://archive.ics.uci.edu/ml/datasets/BitcoinHeistRansomwareAddressDataset}}. Using a time interval of 24 hours, the contributors of this dataset \cite{Akcora2020} extracted daily transactions on the network and formed the Bitcoin graph. In total, we have 2,916,697 observations of six numerical features (the remaining ones were discarded).

Due to the nature of this dataset, we do not observe typical statistical properties (there are no seasonal patterns, the data is not stationary, the fluctuations do not show evident patterns and features are not normally distributed). Table~\ref{tab:stats_bitcoin} depicts descriptive statistics for the retained features.

\begin{table}[!htp]
\centering 
\caption{Descriptive statistics (mean, standard deviation, minimum and maximum value) of variables in the Bitcoin dataset.}
\label{tab:stats_bitcoin}
\begin{tabular}{|c|c|c|c|c|}
\hline
   variable &mean&std	&	min	&max \\\hline
   length  	&45.01&58.98&0.00&	144.00\\\hline
   weight		& 0.55&	3.67&0.00&  1,943.75\\\hline       
   count   &721.64		&	1,689.68&		1.00&	14,497.00	\\\hline
   looped	&238.51	&966.32&0.00&14,496.00\\\hline
   neighbors &		2.21&	17.92&	1.00 &12,920.00\\\hline
   income&4.47e+09 &       	1.63e+11 	&3.00e+07&	5.00e+13\\\hline
\end{tabular}
\end{table}

Table \ref{tab:bitcoin} shows the errors obtained by each optimized network when forecasting 200 steps ahead, and the training and test times (in seconds). After performing hyperparameter tuning, we found that the optimal batch size for RNN, GRU and LSTM was 4,096, 128 and 64, respectively, while the number of LSTCN iterations was set to eight. In this problem, the LSTCN model clearly outperformed the remaining algorithms selected for comparison. When it comes to the training time, LSTCN is 2.1E+1 times faster than RNN, 1.7E+3 times faster than GRU, and 2.2E+3 times faster than LSTM. Similarly to the other experiments, RNN was the second-fastest algorithm; however, it yields the highest forecasting errors.

\begin{table}[!htp]
\centering
\caption{Simulation results for the Bitcoin case study.}
\label{tab:bitcoin}
\begin{tabular}{|c|c|c|c|c|}
\hline
       & \multicolumn{2}{c|}{error} & \multicolumn{2}{c|}{time} \\ \hline
method & training   & test      & training  & test      \\ \hline
RNN    & 0.3003     & 0.3146    & 32.81     & 1.56      \\ \hline
GRU    & 0.0653     & 0.0918    & 2596.92   & 5.07      \\ \hline
LSTM   & 0.0664     & 0.0872    & 3488.91   & 6.96      \\ \hline
LSTCN  & 0.0583     & 0.0774    & 1.56      & 0.09       \\ \hline
\end{tabular}
\end{table}

\begin{figure}[!ht]
\centering
	\includegraphics[width=0.48\textwidth]{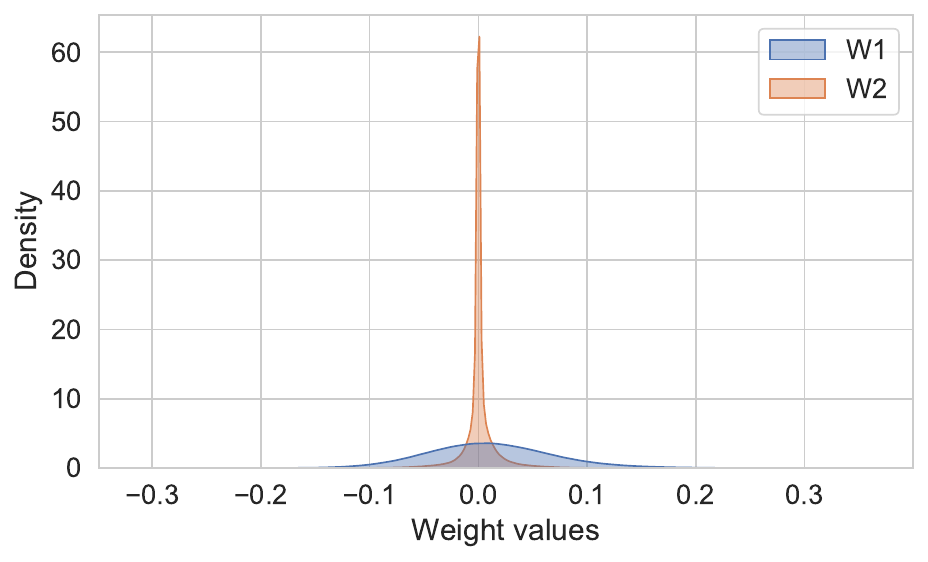}
\caption{Distribution of weights in the Bitcoin case study.}
\label{fig:histogram_bitcoin}
\end{figure}

Figure \ref{fig:histogram_bitcoin} shows the distribution of weights in the first prior knowledge matrix and the matrix computed in the last learning process. This figure illustrates how the weights become more sparse as the network performs more iterations. This happens due to a heavy $\ell_2$ regularization with $\lambda$=1.0E+3 being the best penalty value obtained with grid search. However, no shift in the distributions is observed.

\begin{figure}[!ht]
\centering
	\includegraphics[width=0.48\textwidth]{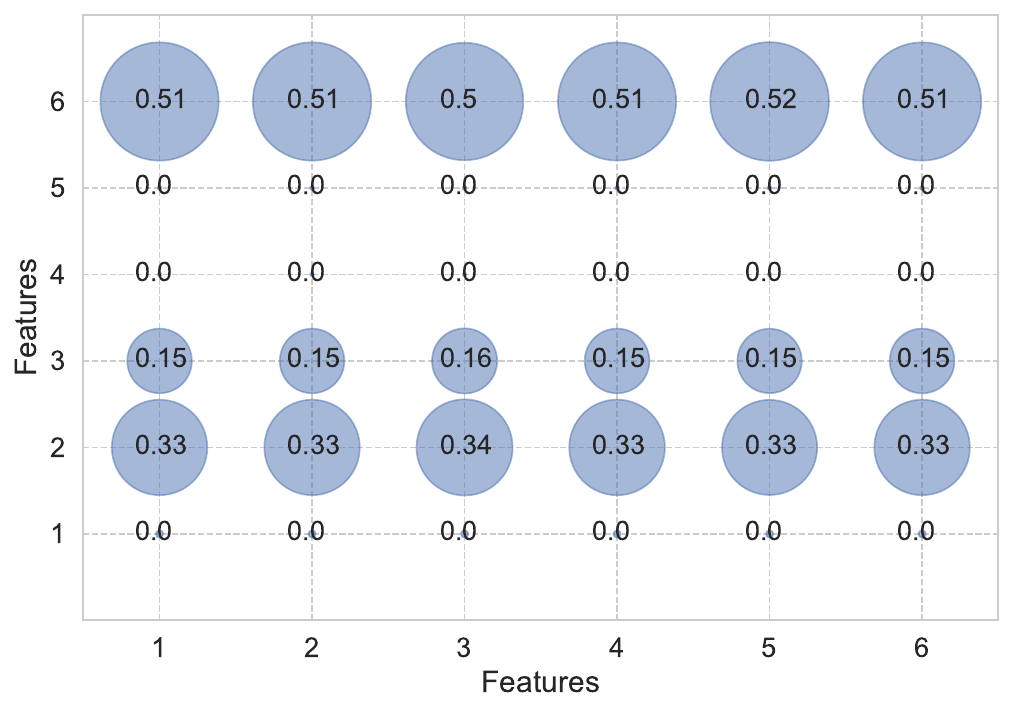}
\caption{Feature influence in the Bitcoin case study.}
\label{fig:relevance_bitcoin}
\end{figure}

Figure \ref{fig:relevance_bitcoin} shows the feature influence scores obtained for the Bitcoin case study. Similarly to the previous scenario, these scores were computed after averaging the $W_1$ and $W_2$ matrices that result from adjusting the network to the last time patch.

The relevance scores suggest that the sixth (income), the second (weight) and the third (count) features have the biggest influence in the forecasting. It is worth mentioning that feature influence scores might not be aligned to the expectations of domain experts. Instead, these scores provide insights into how the network forecasts the next sequence of values.

\section{Concluding remarks}
\label{sec:remarks}

In this paper, we have presented a recurrent neural system termed Long Short-term Cognitive Networks to forecast long time series. The proposed model consists of a collection of STCN blocks, each processing a specific data chunk (time patch). In this neural ensemble, each STCN block passes information to the next one in the form of prior knowledge matrices that preserve what the model has learned up to the current iteration. This means that, in each iteration, the learning problem narrows down to solving a regression problem. Furthermore, neurons and weights can be mapped to the problem domain, making our neural system interpretable.

The numerical simulations using three case studies allow us to draw the following conclusions. Firstly, our model performs better than (or comparably to) state-of-the-art recurrent neural networks. It has not escaped our notice that these algorithms could have produced smaller forecasting errors if we would have optimized other hyperparameters (such as the learning rate, the optimizer, the regularizer, etc.). However, such an increase in performance would come at the expense of a significant increase in the computations needed to produce fully optimized models. Secondly, the simulation results have shown that our proposal is noticeably faster than GRU and LSTM, which are popular recurrent models for time series forecasting. Such a conclusion is particularly relevant since our primary goal was to design a fairly accurate forecasting model with fast training time rather than outperforming the forecasting capabilities of these recurrent models. Finally, we have illustrated how to derive insights into the relevance of features using the network's knowledge structures with little effort.

Future research efforts will be devoted to exploring the forecasting capabilities of our model further. On the one hand, we plan to conduct a larger experiment involving more univariate and multivariate time series. On the other hand, we will analytically study the generalization properties of LSTCNs under the PAC-Learning formalism. This seems especially interesting since the network's size depends on the number of features and the number of steps ahead to be forecast.




\bibliographystyle{elsarticle-harv}

\end{document}